\documentclass{article}
\usepackage{amsmath,graphicx,spconf}
\usepackage{amsthm}
\usepackage[T1]{fontenc}
\usepackage{hyperref, times, cite}
\usepackage{url, datetime}
\usepackage{enumitem,kantlipsum}
\usepackage{mathtools}
\usepackage{bm}
\usepackage{cool}
\usepackage{amssymb,amsfo nts,amsmath,amsthm,amscd,dsfont,mathrsfs}
\usepackage{graphicx,float,psfrag,epsfig,amssymb}
\usepackage{wrapfig}
\usepackage{relsize}
\usepackage{color}
\usepackage{pict2e}
\usepackage{algorithm}
\usepackage{algorithmic}
\usepackage{caption}
\usepackage{nameref}
\usepackage{enumerate}
\usepackage{stackrel}

\usepackage{caption}
\usepackage{subcaption}

\title{Graph Learning from Multivariate Dependent Time Series via a Multi-Attribute Formulation}
\name{{\em Jitendra K.\ Tugnait}  \thanks{This work was supported by NSF Grant ECCS-2040536. Author's email: tugnajk@auburn.edu} \vspace*{-0.1in}}
\address{Department of Electrical \& Computer Engineering\\
Auburn University, Auburn, AL 36849, USA}
\begin{document}
\ninept

\maketitle

\setcounter{footnote}{-1}
\def\thefootnote{\fnsymbol{footnote}}
\renewcommand{\algorithmicrequire}{\textbf{Input:}}
\renewcommand{\algorithmicensure}{\textbf{Output:}}

\begin{abstract}
We consider the problem of inferring the conditional independence graph (CIG) of a high-dimensional stationary multivariate Gaussian time series. In a time series graph, each component of the vector series is represented by distinct node, and associations between components are represented by edges between the corresponding nodes. We formulate the  problem as one of multi-attribute graph estimation for random vectors where a vector is associated with each node of the graph. At each node, the associated random vector consists of a time series component and its delayed copies. We present an alternating direction method of multipliers (ADMM) solution to minimize a sparse-group lasso penalized negative pseudo log-likelihood objective function to estimate the precision matrix of the random vector associated with the entire multi-attribute graph. The time series CIG is then inferred from the estimated precision matrix. A theoretical analysis is provided. Numerical results illustrate the proposed approach which outperforms existing frequency-domain approaches in correctly detecting the graph edges.
\end{abstract}


\section{Introduction} \label{intro}

Graphical models are an important and useful tool for analyzing multivariate data \cite{Lauritzen1996}. Given a collection of random variables, one wishes to assess the relationship between two variables, conditioned on the remaining variables. In graphical models, graphs are used to display the conditional independence structure of the variables. Consider a graph ${\cal G} = \left( V, {\cal E} \right)$ with a set of $p$ vertices (nodes) $V = \{1,2, \cdots , p\} =[p]$, and a corresponding set of (undirected) edges ${\cal E} \subseteq [p] \times [p]$. Also consider a stationary (real-valued), zero-mean,  $p-$dimensional multivariate Gaussian time series ${\bm x}(t)$, $t=0, \pm 1, \pm 2, \cdots $, with $i$th component $x_i(t)$. Given $\{ {\bm x}(t) \}$, in the corresponding graph ${\cal G}$, each component series $\{ x_i(t) \}$ is represented by a node ($i$ in $V$), and associations between components $\{ x_i (t) \}$ and $\{ x_j(t) \}$ are represented by edges between nodes $i$ and $j$ of ${\cal G}$. In a conditional independence graph (CIG), there is no edge between nodes $i$ and $j$ if and only if (iff) $x_i(t)$ and $x_j(t)$ are conditionally independent given the remaining $p$-$2$ scalar series $x_\ell(t)$, $\ell \in [p]$, $\ell \neq i$, $\ell \neq j$ \cite{Dahlhaus2000}. 

Graphical models were originally developed for random vectors (whose statistics are estimated via multiple independent realizations) \cite[p.\ 234]{Eichler2012}. Such models have been extensively studied, and found to be useful in a wide variety of applications \cite{Danaher2014, Friedman2004, Lauritzen2003, Meinshausen2006, Mohan2014}. Graphical modeling of real-valued time-dependent data (stationary time series) originated with \cite{Brillinger1996}, followed by \cite{Dahlhaus2000}. A key insight in \cite{Dahlhaus2000} was to transform the series to the frequency domain and express the graph relationships in the frequency domain. Nonparametric approaches for graphical modeling of real time series in high-dimensional settings ($p$ is large and/or sample size $n$ is of the order of $p$) have been formulated in the form of group-lasso penalized log-likelihood in frequency-domain in \cite{Jung2015a}. Sparse-group lasso penalized log-likelihood approach in frequency-domain has been considered in \cite{Tugnait18c, Tugnait20, Tugnait21b}. 

In this paper we investigate graph structure estimation for stationary Gaussian multivariate time series using a time-domain approach, unlike \cite{Jung2015a, Tugnait18c, Tugnait20} who, as noted earlier, use a frequency-domain approach. After reviewing some graphical modeling background in Sec.\ \ref{GM}, we first reformulate the  problem in Sec.\ \ref{MAMTS} as one of multi-attribute graph estimation for random vectors where a vector is associated with each node of the graph. Then in Sec.\ \ref{SGL} we exploit the results of \cite{Tugnait21a} to provide an alternating direction method of multipliers (ADMM) solution to minimize a sparse-group lasso penalized negative pseudo log-likelihood objective function for multi-attribute graph precision matrix estimation. A theoretical analysis is provided in Sec.\ \ref{TA}. Numerical results in Sec.\ \ref{NE} illustrate the proposed approach.

{\em Notation}: We use ${\bm S} \succeq 0$ and ${\bm S} \succ 0$ to denote that {the symmetric matrix} ${\bm S}$ is positive semi-definite and positive definite, respectively. {For a set $V$, $|V|$ or $\mbox{card}(V)$ denotes its cardinality. $\mathbb{Z}$ is the set of integers. Given ${\bm A} \in \mathbb{R}^{p \times p}$, we use $\phi_{\min }({\bm A})$, $\phi_{\max }({\bm A})$, $|{\bm A}|$ and $\mbox{tr}({\bm A})$ to denote the minimum eigenvalue, maximum eigenvalue, determinant and  trace of ${\bm A}$, respectively. For ${\bm B} \in \mathbb{R}^{p \times q}$, we define  $\|{\bm B}\| = \sqrt{\phi_{\max }({\bm B}^\top  {\bm B})}$, $\|{\bm B}\|_F = \sqrt{\mbox{tr}({\bm B}^\top  {\bm B})}$ and $\|{\bm B}\|_1 = \sum_{i,j} |B_{ij}|$, where $B_{ij}$ is the $(i,j)$-th element of ${\bm B}$ (also denoted by $[{\bm B}]_{ij}$). Given ${\bm A} \in \mathbb{R}^{p \times p}$, ${\bm A}^+ = \mbox{diag}({\bm A})$ is a diagonal matrix with the same diagonal as ${\bm A}$, and  ${\bm A}^- = {\bm A} - {\bm A}^+$ is ${\bm A}$ with all its diagonal elements set to zero. 

\section{Graphical Models} \label{GM}
Here we provide some background material for graphical models for random vectors and for multivariate time series. 

\subsection{Random Vectors} \label{MRV}
Consider a graph ${\cal G} = \left( V, {\cal E} \right)$ with a set of $p$ vertices (nodes) $V = \{1,2, \cdots , p\} =[p]$, and a corresponding set of (undirected) edges ${\cal E} \subseteq V \times V$. Let ${\bm x} = [x_1 \; x_2 \; \cdots \; x_p]^\top \in \mathbb{R}^p$ denote a Gaussian random vector that is zero-mean with covariance ${\bm \Sigma} = E\{ {\bm x} {\bm x}^\top \} \succ {\bm 0}$. The conditional independence relationships among $x_i$'s are encoded in ${\cal E}$ where edge $\{ i,j \}$ between nodes $i$ and $j$ exists if and only if (iff) $x_i$ and $x_j$ are conditionally dependent given the remaining $p$-$2$ variables $x_\ell$, $\ell \in [p]$, $\ell \neq i$, $\ell \neq j$. Let 
\begin{equation} \label{neweq1}
   {\bm x}_{-ij} = \{ x_k \, : \, k \in V \textbackslash \{i_, j \} \}  \in \mathbb{R}^{p-2}
\end{equation}
denote the vector ${\bm x}$ after deleting $x_i$ and $x_j$ from it. Let $\bm{\Omega} = \bm{\Sigma}^{-1}$ denote the precision matrix.  Define
\begin{equation}  \label{neweq2}
   e_{i|-ij} = x_i - E \{ x_i | {\bm x}_{-ij} \} \, , \;\; 
	 e_{j|-ij} = x_j - E \{ x_j | {\bm x}_{-ij} \} \, .
\end{equation}
Then we have the following equivalence \cite{Lauritzen1996}
\begin{equation}  \label{neweq3}
  \{ i,j \} \not\in {\cal E} \; \Leftrightarrow \; \Omega_{ij} = 0 
  \;  \Leftrightarrow \; E\{ e_{i|-ij} e_{j|-ij} \} = 0 \, .
\end{equation}
Note that $E \{ x_i | {\bm x}_{-ij} \}$ is linear in ${\bm x}_{-ij}$ since ${\bm x}$ is zero-mean Gaussian, and furthermore it minimizes the conditional mean-square error 
\begin{equation}  \label{neweq4}
   E \{ x_i | {\bm x}_{-ij} \} = \arg \min_{b} E \{ (x_i-b({\bm x}_{-ij}))^2  |  {\bm x}_{-ij} \} \, .
\end{equation}
Similar comments apply to $E \{ x_j | {\bm x}_{-ij} \}$.

\subsection{Multivariate Time Series} \label{MTS}
Consider stationary Gaussian time series ${\bm x}(t) \in \mathbb{R}^p$, $t \in \mathbb{Z}$, with $E\{ {\bm x}(t) \} = 0$ and ${\bm R}_{xx}( \tau ) = \mathbb{E} \{ {\bm x}(t + \tau) {\bm x}^T(t ) \}$, $\tau \in \mathbb{Z}$. The conditional independence relationships among time series components $\{ x_i(t) \}$'s are encoded in edge set ${\cal E}$ of ${\cal G} = \left( V, {\cal E} \right)$,  $V = [p]$, ${\cal E} \subseteq V \times V$, where edge $\{ i,j \} \not\in {\cal E}$ iff $\{ x_i(t), \, t \in \mathbb{Z} \}$ and $\{ x_j(t), \, t \in \mathbb{Z} \}$ are conditionally independent given the remaining $p$-$2$ components 
\begin{equation} \label{neweq10}
   {\bm x}_{-ij,\mathbb{Z}} = \{ x_k(t) \, : \, k \in V \textbackslash \{i_, j \}, \, 
	    t \in \mathbb{Z}  \}   \, .
\end{equation}
Define
\begin{align}  
   e_{i|-ij}(t) = & x_i(t) - E \{ x_i(t) | {\bm x}_{-ij,\mathbb{Z}} \}  \label{neweq11} \\
	 e_{j|-ij}(t) = & x_j(t) - E \{ x_j(t) | {\bm x}_{-ij,\mathbb{Z}} \} \, , \label{neweq12}
\end{align}
and the power spectral density (PSD) matrix ${\bm S}_{x}(f)$
\begin{equation}  \label{neweq13}
   {\bm S}_x(f) = \sum_{\tau = -\infty}^{\infty}  {\bm R}_{xx}( \tau ) e^{-j 2 \pi f \tau} \, .
\end{equation}
Then we have the following equivalence \cite{Dahlhaus2000}
\begin{align}  
  \{ i,j \} \not\in {\cal E} & \; \Leftrightarrow \; [{\bm S}_x^{-1}(f)]_{ij} = 0 \; 
	              \forall f \in [0,1]  \nonumber \\
  & \;  \Leftrightarrow \; E\{ e_{i|-ij}(t+\tau) e_{j|-ij}(t) \} = 0 \; \forall \tau \in \mathbb{Z} \, .
	       \label{neweq14}
\end{align}

\subsection{Multi-Attribute Graphical Models for Random Vectors} \label{MAMRV}
Now consider $p$ jointly Gaussian vectors ${\bm z}_i \in \mathbb{R}^m$, $i \in [p]$. We associate ${\bm z}_i$ with the $i$th node of graph ${\cal G} = \left( V, {\cal E} \right)$,  $V = [p]$, ${\cal E} \subseteq V \times V$. We now have $m$ attributes per node. Now $\{ i,j \} \not\in {\cal E}$ iff vectors ${\bm z}_i$ and ${\bm z}_j$ are conditionally independent given the remaining $p$-$2$ vectors $\{ {\bm z}_\ell \, , \ell \in V \textbackslash \{i_, j \} \}$. Let
\begin{equation}  \label{neweq20}
   {\bm x} = [ {\bm z}_1^\top \; {\bm z}_2^\top \; \cdots \; {\bm z}_p^\top ]^\top 
	              \in \mathbb{R}^{mp} \, .
\end{equation}
Let ${\bm \Omega} = (E \{ {\bm x} {\bm x}^\top \} )^{-1}$ assuming $E \{ {\bm x} {\bm x}^\top \} \succ {\bm 0}$. Define the $m \times m$ subblock ${\bm \Omega}^{(ij)}$ of ${\bm \Omega}$ as
\begin{equation}  \label{neweq21}
   [{\bm \Omega}^{(ij)}]_{rs} = [{\bm \Omega}]_{(i-1)m+r, (j-1)m+s} \, , \; r,s=1,2, \cdots , m \, .
\end{equation}
Let
\begin{equation} \label{neweq22}
   {\bm z}_{-ij} = \{ {\bm z}_k \, : \, k \in V \textbackslash \{i_, j \} \}  \in \mathbb{R}^{m(p-2)}
\end{equation}
denote the vector ${\bm x}$ in (\ref{neweq20}) after deleting vectors ${\bm z}_i$ and ${\bm z}_j$ from it. Define
\begin{equation}  \label{neweq23}
   {\bm e}_{i|-ij} = {\bm z}_i - E \{ {\bm z}_i | {\bm z}_{-ij} \} \, , \;\; 
	 {\bm e}_{j|-ij} = {\bm z}_j - E \{ {\bm z}_j | {\bm z}_{-ij} \} \, .
\end{equation}
Then we have the following equivalence \cite{Kolar2014}
\begin{equation}  \label{neweq24}
  \{ i,j \} \not\in {\cal E} \; \Leftrightarrow \; {\bm \Omega}^{(ij)} = {\bm 0}
  \;  \Leftrightarrow \; E\{ {\bm e}_{i|-ij} {\bm e}_{j|-ij}^\top \} = {\bm 0} \, ,
\end{equation}
where the first equivalence in (\ref{neweq24}) is given in \cite[Sec.\ 2.1]{Kolar2014} and the second equivalence is given in \cite[Appendix B.3]{Kolar2014}.

\section{Multi-Attribute Formulation for Time Series Graphical Modeling} \label{MAMTS}
Consider time series $\{\bm x(t) \}$ as in Sec.\ \ref{MTS}. For some $d \ge 1$, let
\begin{align}  
   {\bm z}_i(t) = & [ x_i(t) \; x_i(t-1) \; \cdots \; x_i(t-d) ]^\top \in \mathbb{R}^{d+1}
	               \label{neweq30} \\
	{\bm y}(t) = & [ {\bm z}_1^\top (t) \; {\bm z}_2^\top (t) \; \cdots \; 
	         {\bm z}_p^\top(t) ]^\top  \in \mathbb{R}^{(d+1)p}
	               \, .\label{neweq31}
\end{align}
Let ${\bm \Omega}_y = (E\{ {\bm y}(t) {\bm y}^\top(t) \})^{-1}$. With $m=d+1$, define the $m \times m$ subblock ${\bm \Omega}_y^{(ij)}$ of ${\bm \Omega}_y$ as
\begin{equation}  \label{neweq32}
   [{\bm \Omega}_y^{(ij)}]_{rs} = [{\bm \Omega}_y]_{(i-1)m+r, (j-1)m+s} \, , \; s,t=1,2, \cdots , m \, .
\end{equation}
Let
\begin{equation} \label{neweq33}
   {\bm z}_{-ij}(t) = \{ {\bm z}_k(t) \, : \, k \in V \textbackslash \{i_, j \} \} \, ,
\end{equation}
\begin{align}  
   {\bm e}_{i|-ij}(t) = & {\bm z}_i(t) - E \{ {\bm z}_i(t) | {\bm z}_{-ij}(t) \} 
	    \label{neweq34} \\
	 {\bm e}_{j|-ij}(t) = & {\bm z}_j(t) - E \{ {\bm z}_j(t) | {\bm z}_{-ij}(t) \} \, . \label{neweq35}
\end{align}
Then by Sec.\ \ref{MAMRV},
\begin{equation}  \label{neweq36}
  \{ i,j \} \not\in {\cal E} \; \Leftrightarrow \; {\bm \Omega}_y^{(ij)} = {\bm 0} \, .
\end{equation}

Define
\begin{equation} \label{neweq37}
   \tilde{\bm x}_{-ij;t,d} = \{ {\bm x}_k(s) \, : \, k \in V \textbackslash \{i_, j \} \, ,
	     t-d \le s \le t \} \, ,
\end{equation}
\begin{align}  
   e_{xi|-ij}(t') = & x_i(t') - E \{ x_i(t') | \tilde{\bm x}_{-ij;t,d} \} 
	    \label{neweq38} \\
	 e_{xj|-ij}(t') = & x_j(t') - E \{ x_j(t') | \tilde{\bm x}_{-ij;t,d} \} \, . \label{neweq39}
\end{align} 
Notice that $e_{xi|-ij}(t')$ above is an element of ${\bm e}_{i|-ij}(t)$ defined in (\ref{neweq34}) for any $t-d \le t' \le t$. Then by (\ref{neweq24}) and (\ref{neweq36}), we have
\begin{equation}  \label{neweq40}
  {\bm \Omega}_y^{(ij)} = {\bm 0} \; \Leftrightarrow \; 
	 E\{ e_{xi|-ij}(t_1) e_{xj|-ij}(t_2) \} =0, \;\; t-d \le t_1, t_2 \le t.
\end{equation}
It follow from (\ref{neweq40}) that if we let $ d \uparrow \infty$, then checking if ${\bm \Omega}_y^{(ij)} = {\bm 0}$ to ascertain (\ref{neweq36}) becomes a surrogate for checking if the last equivalence in (\ref{neweq14}) holds true for time series graph structure estimation without using frequency-domain methods.

\section{Sparse-Group Graphical Lasso Solution to Multi-Attribute Formulation} \label{SGL}
We now consider a finite set of data comprised of $n$ zero-mean observations  ${\bm x}(t)$, $t=0, 1,2, \cdots , n-1$. Pick $d > 1$ and as in (\ref{neweq31}), construct ${\bm y}(t)$ for $t=d, d+1, \cdots , n-1$ with sample size $\bar{n} = n-d$. Define the sample covariance $\hat{\bm{\Sigma}}_y = \frac{1}{\bar{n}} \sum_{t=d}^{n-1} {\bm y}(t) {\bm y}^\top(t)$. If the vector sequence $\{ {\bm y}(t) \}_{t=d}^{n-1}$ were i.i.d., the log-likelihood (up to some constants) would be given by $\ln (|\bm{\Omega}_y|)- {\rm tr} ( \hat{\bm{\Sigma}}_y \bm{\Omega}_y )$ \cite{Tugnait21a}. In our case the sequence is not i.i.d., but we will still use this expression as a pseudo log-likelihood and following \cite{Tugnait21a}, consider the penalized negative pseudo log-likelihood 
\begin{align}  
   &L_{SGL} ({\bm \Omega}_y)  = - \ln (|\bm{\Omega}_y|) 
		        + {\rm tr} ( \hat{\bm{\Sigma}}_y \bm{\Omega}_y  ) + P(\bm{\Omega}_y) , \label{eqth2_30} \\
	 &  P(\bm{\Omega}_y)   =   \alpha \lambda \, \| \bm{\Omega}_y^- \|_1  
	    + (1-\alpha) \lambda \sum_{j \ne k}^p\| \bm{\Omega}_y^{(jk)} \|_F \, , \label{eqth2_30c}
\end{align} 
where $P(\bm{\Omega}_y)$ is a sparse-group lasso penalty \cite{Danaher2014,Tugnait21a,Friedman2010a,Friedman2010b}, with group lasso penalty $(1-\alpha) \lambda \sum_{j \ne k}^p\| \bm{\Omega}_y^{(jk)} \|_F $, $\lambda > 0$ and lasso penalty $\alpha \lambda \, \| \bm{\Omega}_y^- \|_1$, $\lambda > 0$ is a tuning parameter, and $0 \le \alpha \le 1$ yields a convex combination of lasso and group lasso penalties. The function $L_{SGL} ({\bm \Omega}_y)$ is strictly convex in ${\bm \Omega}_y \succ {\bm 0}$.

As in \cite{Tugnait21a}, we use the ADMM approach \cite{Boyd2010} with variable splitting. Using variable splitting, consider
\begin{align} 
 \min_{\bm{\Omega}_y \succ {\bm 0}, {\bm W} } & \Big\{  {\rm tr} ( \hat{\bm{\Sigma}}_y \bm{\Omega}_y  ) 
          -   \ln(|\bm{\Omega}_y|) + P({\bm W})   \Big\}
					 \mbox{ subject to  }  \bm{\Omega}_y = {\bm W}  \, .    \label{eqth2_4000}   
\end{align}
The scaled augmented Lagrangian for this problem is \cite{Boyd2010}
\begin{align} 
 L_\rho = {\rm tr} ( \hat{\bm{\Sigma}}_y \bm{\Omega}_y  ) 
          -   \ln(|\bm{\Omega}_y|) + P({\bm W})   
					 + \frac{\rho}{2}  \| \bm{\Omega}_y - {\bm W} + {\bm U}\|^2_F   \label{eqth2_4010}  
\end{align}
where ${\bm U}$ is the dual variable, and $\rho >0$ is the penalty parameter. Given the results $ \bm{\Omega}^{(i)}, {\bm W}^{(i)}, {\bm U}^{(i)}$ of the $i$th iteration, in the $(i+1)$st iteration, an ADMM algorithm executes the following three updates:
\begin{itemize}
\item[(a)] $\bm{\Omega}_y^{(i+1)} \leftarrow \arg \min_{\bm{\Omega}_y} \, L_a(\bm{\Omega}_y) ,\;\;
           L_a(\bm{\Omega}_y) := {\rm tr} ( \hat{\bm{\Sigma}}_y \bm{\Omega}_y  ) 
          -   \ln(|\bm{\Omega}_y|) + \frac{\rho}{2}  \| \bm{\Omega}_y - {\bm W}^{(i)} + {\bm U}^{(i)}\|^2_F $
\item[(b)] ${\bm W}^{(i+1)}  \leftarrow \arg \min _{ {\bm W} } L_b({\bm W}) , \;\;
          L_b({\bm W}) :=  \alpha \lambda \, \| {\bm{W}}^- \|_1  
	    + (1-\alpha) \lambda \sum_{i \ne j}^p\| {\bm{W}}^{(ij)} \|_F  
					+ \frac{\rho}{2}  \| \bm{\Omega}_y^{(i+1)} - {\bm W} + {\bm U}^{(i)} \|^2_F$
\item[(c)] ${\bm U}^{(i+1)} \leftarrow {\bm U}^{(i)}  +
   \left( \bm{\Omega}_y^{(i+1)} - {\bm W}^{(i+1)} \right)$
\end{itemize}

{\textbf Remark 1}. We follow the detailed ADMM algorithm given in \cite{Tugnait21a} for the above updates; details may be found therein (where we need to replace ${\bm \Omega}$ with ${\bm \Omega}_y$). The parameter tuning (selection of $\lambda$ and $\alpha$) approach given in \cite{Tugnait21a} does not apply (strictly speaking) in our case since our $\{ {\bm y}(t) \}$ is not an i.i.d.\ sequence. $\quad \Box$

\section{Theoretical Analysis} \label{TA}  

In this section we analyze consistency (Theorem 1) by invoking some results from \cite{Tugnait21a}. The difference from \cite{Tugnait21a} is that while the observations in \cite{Tugnait21a} are i.i.d., here $\{ {\bm x}(t) \}$, and $\{ {\bm y}(t) \}$ constructed from it, are dependent sequences. Therefore, we need a model for this dependence. This influences concentration inequality regarding convergence of  sample covariance $\hat{\bm \Sigma}$. Once this aspect is accounted for, \cite[Theorem 1]{Tugnait21a} applies immediately.   

To quantify the dependence structure of $\{ {\bm x}(t) \}$, we will follow \cite{Chen16}; other possibilities include \cite{Basu15, Shu19}. 
\begin{itemize}
\setlength{\itemindent}{0.1in}
\item[(A0)] Assume $\{ {\bm x}(t) \}$ obeys 
\begin{equation}  \label{neweq100}
  {\bm x}(t) = \sum_{i=0}^\infty {\bm A}_i {\bm e}(t-i) \, , 
\end{equation}
where $\{ {\bm e}(t) \}$ is i.i.d., Gaussian, zero-mean with identity covariance, ${\bm e}(t) \in \mathbb{R}^p$, $\, {\bm A}_i \in \mathbb{R}^{p \times p}$, and 
\begin{equation}  \label{neweq101}
  \max_{1 \le q \le p} \sqrt{ \sum_{k=1}^p ([{\bm A}_i]_{qk})^2 } \le \frac{c_a}{(\max(1,i))^\gamma}
\end{equation}
for all $i \ge 0$, some $c_a \in (0, \infty)$, and $\gamma > 1$.
\end{itemize}

Assumption (A0) is satisfied if ${\bm x}(t)$ is generated by an asymptotically stable vector ARMA (autoregressive moving average) model with distinct ``poles,'' satisfying ${\bm x}(t) = -\sum_{i=1}^{q} {\bm \Phi}_i {\bm x}(t-i) +\sum_{i=0}^{r} {\bm \Psi}_i {\bm e}(t-i)$, because in that case $\|{\bm A}_i\|_F \le a |\lambda_0|^i$ for some $0 < a < \infty$ where $|\lambda_0| < 1$ is the largest magnitude ``pole'' (root of $c(z) := \big|{\bm I} + \sum_{i=1}^{q} {\bm \Phi}_i z^{-i} \big| =0$) of the model. It can be shown that there exist $0 < b < \infty$ and $1 < \gamma < \infty$ such that $ a |\lambda_0|^i \le b \, i^{-\gamma}$ for $i \ge 1$, thereby satisfying assumption (A0).
 
By Assumption (A0), it follows that ${\bm y}(t) = \sum_{i=0}^\infty {\bm B}_i \bar{\bm e}(t-i)$, $\bar{\bm e}(t) \in \mathbb{R}^{mp}$ is i.i.d., Gaussian, zero-mean with identity covariance, $m=d+1$, $\, {\bm B}_i \in \mathbb{R}^{(mp) \times (mp)}$, for some ${\bm B}_i$'s such that
\begin{equation}  \label{neweq102}
  \max_{1 \le q \le mp} \sqrt{ \sum_{k=1}^{mp} ([{\bm B}_i]_{qk})^2 } \le \frac{{c}_a}{(\max(1,i))^\gamma}
\end{equation}
for all $i \ge 0$, with ${c}_a$, and $\gamma$ as in Assumption (A0).
Then we have Lemma 1, following \cite[Lemma VI.2, supplementary]{Chen16} for the case $\gamma > 1$ ($\gamma$ is called $\beta$ in \cite{Chen16}). \\
{\it Lemma 1}: Under Assumption (A0), the sample covariance $\hat{\bm{\Sigma}}_y$ satisfies the tail bound
\begin{equation}  \label{naeq58bx}
   P \left(   \Big| [ \hat{\bm{\Sigma}}_y - \bm{\Sigma}_{y0} ]_{kl} \Big| 
	    \ge \delta \right) \le 2 \, \exp(-C_u \bar{n} \min(\delta^2, \delta)) 
\end{equation}
for any $\delta > 0$ and for any $k,l$, where $C_u \in (0, \infty)$ is an absolute (universal) constant. $\quad \bullet$ \\
Constant $C_u$ results from the application of the Hanson-Wright inequality \cite{Rudelson13}.

In rest of this section we allow $p$ and $\lambda$ to be a functions of sample size $n$, denoted as $p_n$ and $\lambda_n$, respectively. Lemma 1 leads to Lemma 2. \\
{\it Lemma 2}: Under Assumption (A0), the sample covariance $\hat{\bm{\Sigma}}_y$ satisfies the tail bound
\begin{equation}  \label{naeq58bx}
   P \left(  \max_{k,l} \Big| [ \hat{\bm{\Sigma}}_y - \bm{\Sigma}_{y0} ]_{kl} \Big|
	    > C_0 \sqrt{\frac{\ln(m p_n)}{\bar{n}}} \right) \le \frac{1}{(mp_n)^{\tau -2}} 
\end{equation}
for $\tau > 2$, if the sample size  $\bar{n}=n-d >  N_1 = \ln(2 (mp_n)^\tau )/C_u$, where $m=d+1$ and $C_0 = \sqrt{N_1/\ln(mp_n)}$. $\quad \bullet$

Lemma 2 above replaces \cite[Lemma 2]{Tugnait21a} for dependency in observations. Further assume
\begin{itemize}
\setlength{\itemindent}{0.1in}
\item[(A1)] Let $\bm{\Sigma}_{y0} = E\{ {\bm y}(t) {\bm y}^\top (t) \} \succ {\bm 0}$ denote the true covariance of ${\bm y}(t)$. Define ${\cal E}_{y0} = \{ \{i,j\} ~:~ {\bm \Omega}_{y0}^{(ij)} \ne {\bm 0}, ~i\ne j \}$ where $\bm{\Omega}_{y0} = \bm{\Sigma}_{y0}^{-1}$. Assume that card$({\cal E}_{y0}) =|({\cal E}_0)| \le s_{n0}$.

\item[(A2)] The minimum and maximum eigenvalues of $\bm{\Sigma}_{y0}$ satisfy 
\[
     0 < \beta_{\min} \le \phi_{\min}(\bm{\Sigma_{y0}}) \le \phi_{\max}(\bm{\Sigma_{y0}}) \le \beta_{\max} < \infty \, .
\]
Here $\beta_{\min}$ and $\beta_{\max}$ are not functions of $n$.
\end{itemize}

Let $\hat{\bm{\Omega}}_{y\lambda} = \arg\min_{\bm{\Omega}_y \succ {\bm 0}}  L_{SGL} ({\bm \Omega}_y)$.
Theorem 1 establishes consistency of $\hat{\bm{\Omega}}_{y\lambda}$ and it follows by replacing \cite[Lemma 2]{Tugnait21a} with Lemma 2 of this paper in the proof of \cite[Theorem 1]{Tugnait21a}. \\
{\it Theorem 1 (Consistency)}: For $\tau > 2$, let $m=d+1$ and
\begin{equation}  \label{naeq58}
   C_0 = \sqrt{ \ln(2 (mp_n)^\tau ) / ( C_u \ln (m p_n) ) } \, .
\end{equation}
Given real numbers $\delta_1 \in (0,1)$, $\delta_2 > 0$ and $C_1 > 0$, let $C_2 = \sqrt{m}+1+C_1$, and
\begin{align}  
    M = & (1+\delta_1)^2 (2C_2 + \delta_2) C_0 / \beta_{\min}^2 ,  \label{neq15ab0} \\
    r_n = & \sqrt{ \frac{(m p_n+ m^2s_{n0}) \ln (mp_n)}{\bar{n}}} = o(1)\, , \label{neq15ab1} \\
    N_1 = &  \ln(2 (mp_n)^\tau )/C_u ,  \label{neq15ab2} \\
		N_2 = &  \arg \min \left\{ \bar{n} \, : \, r_n \le 
		    \frac{ \delta_1 \beta_{\min}}{ (1+\delta_1)^2 (2C_2 + \delta_2) C_0 } \right\} \, . \label{neq15ab3}
\end{align}
Suppose the regularization parameter $\lambda_n$ and $\alpha \in [0,1]$ satisfy 
\begin{align}  
     \frac{C_1  C_0 }{1+\alpha (m-1)} \sqrt{ \Big( 1+\frac{p_n}{m s_{n0}} \Big) 
		    \frac{\ln (m p_n)}{\bar{n}} } & \ge \frac{\lambda_n}{m} \nonumber \\
				& \hspace*{-0.3in} \ge C_0  \sqrt{\frac{\ln (mp_n)}{\bar{n}}} 
		    \, . \label{neq15abc}
\end{align}
Then if the sample size $\bar{n} = n-d >  \max \{ N_1, N_2 \}$ and assumptions (A0)-(A2) hold true,
$\hat{\bm{\Omega}}_{y\lambda}$ satisfies
\begin{equation}  \label{neq15}
  \| \hat{\bm{\Omega}}_{y\lambda} - \bm{\Omega}_{y0} \|_F 
	        \le M  r_n
\end{equation}
with probability greater than $1-1/(mp_n)^{\tau-2}$. In terms of rate of convergence,  
  $\| \hat{\bm{\Omega}}_{y\lambda} - \bm{\Omega}_{y0} \|_F 
	        = {\cal O}_P \left( r_n \right) $ $\quad \bullet$

\vspace*{-0.07in}
\begin{figure}[ht]
\includegraphics[width=0.9\linewidth]{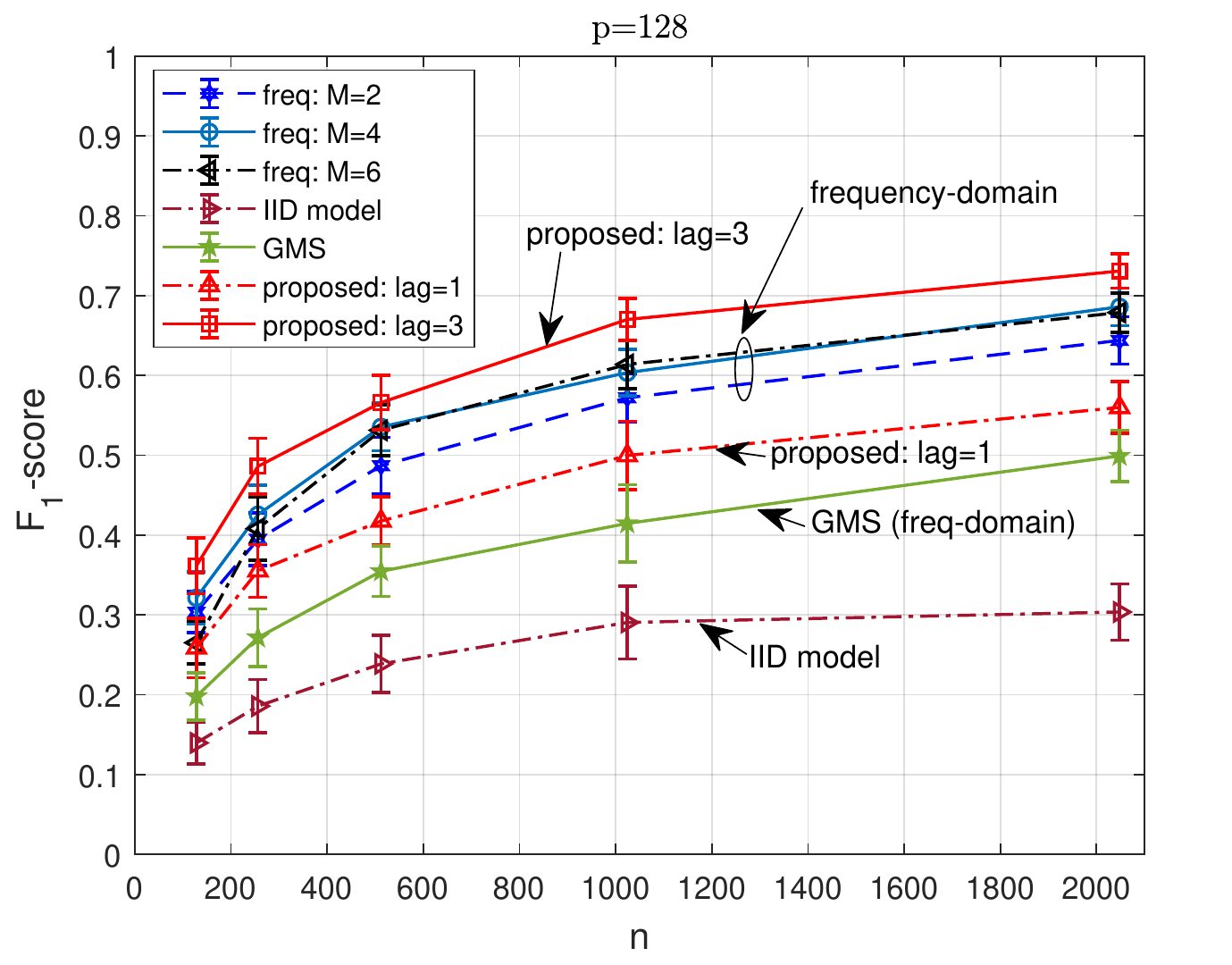} 
\vspace*{-0.1in}
\caption{$F_1$-scores based on 100 runs for 4 approaches. In the proposed approach lag=3 refers to $d=3$. IID model may be viewed as proposed approach with lag=$d=0$.}
\label{fig1}
\end{figure}

\vspace*{-0.2in}
\section{Numerical Example} \label{NE}
Consider $p=128$, 16 clusters (communities) of 8 nodes each, where nodes within a community are not connected to any nodes in other communities. Within any community of 8 nodes, the data are generated using a vector autoregressive (VAR) model of order 3. Consider community $q$, $q=1,2, \cdots , 16$. Then  ${\bm x}^{(q)}(t) \in \mathbb{R}^8$ is generated as
\[
    {\bm x}^{(q)}(t) = \sum_{i=1}^3 {\bm A}^{(q)}_i {\bm x}^{(q)}(t-i) + {\bm w}^{(q)}(t) 
\] 
with  ${\bm w}^{(q)}(t)$ as i.i.d.\ zero-mean Gaussian with identity covariance matrix. Only 10\% of entries of ${\bm A}^{(q)}_i$'s are nonzero and the nonzero elements are independently and uniformly distributed over $[-0.8,0.8]$. We then check if the VAR(3) model is stable with all eigenvalues of the companion matrix $\le 0.95$ in magnitude; if not, we re-draw randomly till this condition is fulfilled. The overall data ${\bm x}(t)$ is given by ${\bm x}(t) = [\, {\bm x}^{(1) \top}(t) \; \cdots \; {\bm x}^{(16) \top}(t) \, ]^\top \in \mathbb{R}^{p}$. First 100 samples are discarded to eliminate transients. This set-up leads to approximately 3.5\% connected edges. The true edge set ${\cal E}_0$ for the time series graph is determined as follows. In each run, we calculated the true PSD ${\bm S}(f)$ for $f \in [0,0.5]$ at intervals of 0.01, and then take $\{i,j\} \in {\cal E}_0$ if $\sum_f | S_{ij}^{-1}(f) | > 10^{-6}$, else $\{i,j\} \not\in {\cal E}_0$.  

Simulation results based on 100 runs are shown in Figs.\ \ref{fig1} and \ref{fig2}. 
The performance measure is $F_1$-score for efficacy in edge detection. The $F_1$-score is defined as $F_1 = 2 \times \mbox{precision} \times \mbox{recall}/(\mbox{precision} + \mbox{recall})$ where $\mbox{precision} = | \hat{\cal E} \cap {\cal E}_0|/ |\hat{\cal E}|$, $\mbox{recall} = |\hat{\cal E} \cap {\cal E}_0|/ |{\cal E}_0|$, and ${\cal E}_0$ and $ \hat{\cal E}$ denote the true and estimated edge sets, respectively. 
Four approaches were tested: \textcolor{red}{{\bf (i)} Proposed multi-attribute graph based approach} with lags (delays) $d=1$ or $d=3$, labeled ``proposed, lag=1'' or ``proposed, lag=3'' in the figures. \textcolor{red}{{\bf (ii)} Frequency-domain sparse-group lasso approach} of \cite{Tugnait18c, Tugnait20, Tugnait21b}, optimized using ADMM, using varying number $M$ (=2,4 or 6) of smoothed PSD estimators in frequency range (0,0.5), labeled ``freq: M=2'', ``freq: M=4'' ``freq: M=6''.  \textcolor{red}{{\bf (iii)} An i.i.d.\ modeling approach} that exploits only the sample covariance $\frac{1}{n} \sum_{t=0}^{n-1} {\bm x}(t) {\bm x}^\top(t)$ (labeled ``IID model''), implemented via the ADMM lasso approach (\cite[Sec.\ 6.4]{Boyd2010}). In this approach, as discussed in Sec.\ \ref{MRV},  edge $\{i,j\}$ exists in the CIG iff $\Omega_{ij} \ne 0$ where precision matrix ${\bm \Omega} = {\bm R}_{xx}^{-1}(0)$. \textcolor{red}{{\bf (iv)} The frequency-domain ADMM approach of \cite{Jung2015a}}, labeled ``GMS'' (graphical model selection), which was applied with $F=4$ (four frequency points, corresponds to $M=4$ in \cite{Tugnait18c, Tugnait20, Tugnait21b}) and all other default settings of \cite{Jung2015a} to compute the PSDs. The tuning parameters, $(\alpha, \lambda)$ for proposed and frequency-domain sparse-group lasso approach of\cite{Tugnait18c, Tugnait20, Tugnait21b}, and lasso parameter $\lambda$ for IID and GMS, were selected via an exhaustive search over a grid of values to maximize the $F_1$-score (which requires knowledge of the true edge-set). The results shown in Figs.\ \ref{fig1} and \ref{fig2} are based on these optimized tuning parameters. (In practice, one would use an information criterion or cross-validation to select the tuning parameters.)

The $F_1$-scores are shown in Fig.\ \ref{fig1} and average timings per run are shown in Fig.\ \ref{fig2} for sample sizes $n=128,256,512,1024, 2048$. It is seen that with $F_1$-score as the performance metric, our proposed method with lag $d=3$ significantly outperforms other approaches while also being faster than frequency-domain approaches.

\vspace*{-0.15in}
\begin{figure}[ht]
  \includegraphics[width=0.9\linewidth]{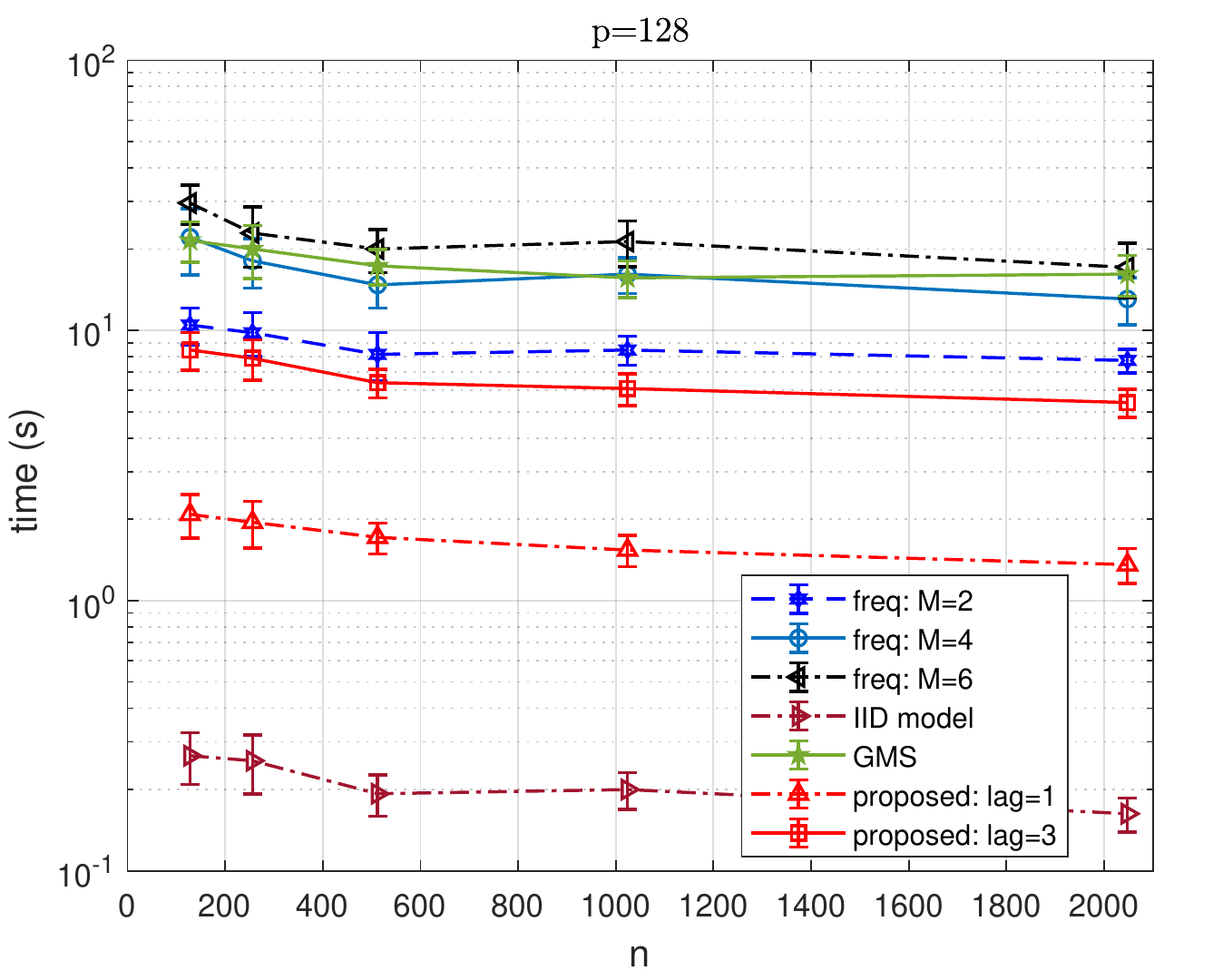} 
	\vspace*{-0.12in}
  \caption{Average timing per run based on 100 runs for 4 approaches.}
\label{fig2}
\end{figure}

\vspace*{-0.25in}
\section{Conclusions} Graphical modeling of dependent Gaussian time series was considered. We formulated the  problem as one of multi-attribute graph estimation for random vectors where a vector is associated with each node of the graph. At each node, the associated random vector consists of a time series component and its delayed copies. We exploited the results of \cite{Tugnait21a} to provide an ADMM solution to minimize a sparse-group lasso penalized negative pseudo log-likelihood objective function for multi-attribute graph precision matrix estimation. A theoretical analysis was provided. Numerical results were provided to illustrate the proposed approach which outperforms the approaches of \cite{Jung2015a, Tugnait18c, Tugnait20, Tugnait21b} with $F_1$-score as the performance metric for graph edge detection.

\bibliographystyle{unsrt} 

\end{document}